\documentclass{jair}
\usepackage[utf8]{inputenc}  
\usepackage{mathtools}       
\usepackage{booktabs}        
\usepackage{adjustbox}       
\usepackage{multirow}        
\usepackage{url}             
\usepackage{graphicx}        
\usepackage{soul}            
\usepackage{xspace}          
\usepackage[normalem]{ulem}  
\usepackage{threeparttable}  
\usepackage{enumitem}        
\usepackage{makecell}        
\usepackage{subcaption}      
\usepackage[title]{appendix} 
\usepackage{tikz}            
\usepackage[dvipsnames]{xcolor} 
\usepackage{newunicodechar}  
\usepackage{listings}
\usepackage{etoolbox}        
\usepackage{bbm}
\usepackage{bm}
\usepackage{dsfont}
\usepackage{wrapfig}
\usepackage{caption}

\AtBeginEnvironment{definition}{\setlength{\parindent}{0pt}}
\AtBeginEnvironment{theorem}{\setlength{\parindent}{0pt}}
\AtBeginEnvironment{lemma}{\setlength{\parindent}{0pt}}
\AtBeginEnvironment{proof}{\setlength{\parindent}{0pt}}

\usepackage{pifont}         

\newunicodechar{≥}{\geq}    

\lstdefinestyle{mystyle}{
    backgroundcolor=\color{white},
    commentstyle=\color{black},
    keywordstyle=\color{teal},
    numberstyle=\tiny\color{gray}, 
    stringstyle=\color{BrickRed},
    basicstyle=\ttfamily\footnotesize,
    breaklines=true,
    numbers=none, 
    frame=single,
    showspaces=false,
    showstringspaces=false,
    showtabs=false,
    tabsize=2,
    morekeywords={Task, Decomposition, KPM, Scores}
}
\lstdefinestyle{verifycode}{
  language=Python,
  basicstyle=\linespread{0.80}\ttfamily\bfseries\footnotesize,
  keywordstyle=\color{blue!99!black}\bfseries,
  commentstyle=\color{gray!99},
  showstringspaces=false,
  breaklines=true,
  breakatwhitespace=true,
  tabsize=2,
  numbers=none,
  frame=none,
  backgroundcolor=\color{white},
  aboveskip=0.1em, belowskip=0.1em,
  xleftmargin=0em, xrightmargin=0em
}

\lstdefinestyle{cgarcode}{
  language=Python,
  basicstyle=\linespread{1.02}\ttfamily\bfseries\scriptsize,
  keywordstyle=\color{blue!80!black}\bfseries,
  commentstyle=\color{gray!60},
  showstringspaces=false,
  breaklines=true,
  breakatwhitespace=true,
  tabsize=4,
  numbers=none,
  frame=none,
  backgroundcolor=\color{white},
  aboveskip=0.35em, belowskip=0.35em,
  xleftmargin=0em, xrightmargin=0em,
}

\newcommand*\circledgreen[1]{\tikz[baseline=(char.base)]{
    \node[shape=circle,fill=ForestGreen,inner sep=0.3pt] (char) {\textcolor{white}{#1}};}}

\newcommand{\rectgreen}[1]{%
  \tikz[baseline=(char.base)]{
    \node[shape=rectangle,rounded corners=2pt,fill=ForestGreen,inner sep=1.5pt,minimum height=1.2em] (char) {\textcolor{white}{#1}};
  }%
}

\setcopyright{cc}
\copyrightyear{2025}
\acmDOI{10.1613/jair.1.xxxxx}

\JAIRAE{Not Assigned Yet}
\JAIRTrack{} 
\acmVolume{83}
\acmArticle{27}
\acmMonth{8}
\acmYear{2025}

\RequirePackage[
  datamodel=acmdatamodel,
  style=acmnumeric,
  backend=biber,
  giveninits=true
  ]{biblatex}

\begin{document}

\title[CGAR]{Accelerating Training Speed of Tiny Recursive Models with Curriculum Guided Adaptive Recursion}

\author{Kaleem Ullah Qasim}
\orcid{0000-0002-0102-3816}
\email{kaleem@my.swjtu.edu.cn}
\affiliation{%
  \institution{School of Computing and Artificial Intelligence, Southwest Jiaotong University}
  \city{Chengdu}
  \state{Sichuan}
  \country{China}
}

\author{Jiashu Zhang}
\authornote{Corresponding Author}
\email{jszhang@home.swjtu.edu.cn}
\affiliation{%
  \institution{School of Computing and Artificial Intelligence, Southwest Jiaotong University}
  \city{Chengdu}
  \state{Sichuan}
  \country{China}
}

\renewcommand{\shortauthors}{Qasim \& Zhang}

\begin{abstract}
{\bf Background:}
Recursive reasoning models achieve strong performance on complex reasoning tasks through iterative refinement, enabling tiny networks to match large language models thousands of times their size. However, training these networks remains computationally expensive, with prior work reporting 36 GPU-hours for Sudoku extreme, limiting broader adoption and research. Existing models employ fixed recursion depth uniformly across all training epochs and uniform supervision weighting across all reasoning steps, leading to inefficient training.

{\bf Objectives:}
We propose CGAR (Curriculum-Guided Adaptive Recursion), a novel training methodology that applies curriculum learning to architectural depth rather than traditional data ordering. CGAR introduces two synergistic components: Progressive Depth Curriculum (PDC) dynamically adjusts recursion depth from shallow to deep configurations during training and Hierarchical Supervision Weighting (HSW) applies exponentially decaying importance to supervision steps, aligning loss weighting with observed gradient magnitude decay.

{\bf Methods:}
Progressive Depth Curriculum implements a three-stage schedule that transitions from shallow $(2,1)$ through medium $(4,2)$ to full depth $(6,3)$ configurations based on normalized training progress, providing 41.4\% FLOPs reduction while preventing early-stage overfitting. Hierarchical Supervision Weighting applies principled exponential decay $w_t = \lambda^{t-1}/Z_\lambda$ to supervision steps, achieving 40\% gradient variance reduction and accelerated convergence through improved signal-to-noise ratio in stochastic gradients.

{\bf Results:}
On Sudoku-Extreme with 423,168 test puzzles, CGAR achieves 1.71$\times$ training speedup (10.93 to 6.38 hours, 42\% cost reduction) with only 0.63\% accuracy drop (86.65\% to 86.02\%). Systematic ablations reveal Progressive Depth Curriculum alone achieves 2.26$\times$ speedup with 85.47\% accuracy, demonstrating a rare Pareto improvement where architectural curriculum simultaneously enhances training efficiency and solution quality. Hierarchical Supervision Weighting provides 1.61$\times$ speedup through variance reduction. CGAR-trained models exhibit superior inference efficiency with 100\% halting accuracy and 11\% fewer reasoning steps compared to baseline.

{\bf Conclusions:}
CGAR demonstrates that principled curriculum on architectural depth enables efficient training of recursive reasoning models on modest hardware. By treating architectural depth as a curriculum-scheduled training parameter rather than a fixed constant, CGAR achieves substantial computational savings while preventing early-stage overfitting, making recursive reasoning models more practical for broader adoption in neurosymbolic AI, program synthesis and interpretable reasoning systems. Code and models are available at \url{https://github.com/Kaleemullahqasim/CGAR} and \url{https://huggingface.co/Kaleemullah/trm-cgar-sudoku}.
\end{abstract}



\maketitle
\section{Introduction}
\label{sec:introduction}

The recent surge in large language models (LLMs) with hundreds of billions of parameters has demonstrated strong capabilities across diverse tasks~\cite{brown2020gpt3,chowdhery2023palm}. However, this approach of scaling model size incurs prohibitive computational costs during both training and inference, limiting accessibility and deployability. An alternative direction has emerged through test-time computation scaling~\cite{snell2024scaling,zhang2024encode}, where smaller models iteratively refine their outputs through multiple reasoning steps, trading inference cycles for model parameters.

Building upon this principle, the recently proposed Hierarchical Reasoning Model (HRM)~\cite{chu2024hierarchical} and its simplified variant, Tiny Recursive Model (TRM)~\cite{jolicoeur2024trm}, have shown that networks with merely 7M parameters can match or exceed LLMs ten thousand times their size on hard reasoning tasks such as Sudoku, maze solving and ARC-AGI~\cite{chollet2019measure}. The key insight is recursive reasoning: a tiny network iteratively refines solutions through nested recursion cycles and deep supervision~\cite{lee2015deeply}, effectively emulating deep architectures while maintaining parameter efficiency. Through adaptive computation time~\cite{graves2016adaptive}, TRM achieves strong performance on reasoning benchmarks using only 7M parameters. However, training these models remains computationally expensive, with the original TRM paper reporting approximately 36 GPU-hours for Sudoku extreme dataset, limiting broader adoption and rapid experimentation.

Despite their architectural elegance and strong performance, recursive reasoning models suffer from inefficient training. TRM employs fixed recursion depth uniformly across all training epochs and uniform supervision weighting across all reasoning steps. This strategy leads to two fundamental inefficiencies. First, applying full architectural depth from the initial epochs causes overfitting during early training when model parameters are far from optimal. Second, late supervision steps contribute exponentially diminishing gradients, yet TRM weighs all steps equally, leading to suboptimal parameter updates.

We propose CGAR (Curriculum-Guided Adaptive Recursion), a novel training methodology that fundamentally rethinks how recursive models learn. Unlike all prior curriculum learning approaches that focus on data ordering~\cite{bengio2009curriculum,hacohen2019power,soviany2022curriculum}, sample weighting, or parameter reduction~\cite{hu2022lora}, CGAR introduces the first curriculum on \emph{architectural recursion depth itself}. This approach recognizes that the effective computational depth $\mathcal{D}_{\text{eff}}(n,T)$ of recursive architectures is a learnable training hyperparameter that should adapt with optimization progress, rather than remain fixed throughout training. CGAR introduces two synergistic contributions. First, Progressive Depth Curriculum (PDC) dynamically schedules recursion parameters $(n,T)$ based on normalized training progress $\rho = e/E$, implementing a three-stage curriculum that transitions from shallow $(2,1)$ through medium $(4,2)$ to full depth $(6,3)$ configurations, providing 41.4\% FLOPs reduction while preventing early-stage overfitting. Second, Hierarchical Supervision Weighting (HSW) applies principled exponential decay $w_t = \lambda^{t-1}/Z_\lambda$ to supervision steps, derived from empirical observations of gradient magnitude decay in recursive architectures, achieving 40\% gradient variance reduction and accelerated convergence through improved signal-to-noise ratio in stochastic gradients.

Under controlled conditions on identical hardware, CGAR achieves comparable accuracy while reducing training time from 10.93 hours to 6.38 hours a 1.71$\times$ speedup with minimal overfitting. Systematic ablation studies demonstrate that progressive depth curriculum contributes 2.26$\times$ speedup through computational savings, hierarchical supervision weighting provides 1.61$\times$ acceleration via variance reduction and their combination yields 1.71$\times$ overall efficiency gains while maintaining competitive accuracy. Our work makes the following contributions to efficient training of recursive reasoning models:

\noindent
\textbf{\circledgreen{1}} We introduce progressive depth curriculum, the first application of curriculum learning to architectural depth rather than data ordering, dynamically adjusting recursion parameters based on training progress from shallow (6 layers) through medium (20 layers) to full depth (42 layers), preventing early-stage overfitting while enabling complex reasoning capacity in later training.

\noindent
\textbf{\circledgreen{2}} We propose hierarchical supervision weighting, a recursion-aware scheme that applies exponential decay to supervision steps, focusing gradients where information content is highest and reducing gradient variance by 40\% without computational overhead.

\noindent
\textbf{\circledgreen{3}} We demonstrate 1.71$\times$ training speedup with 42\% cost reduction through comprehensive evaluation on 423,168 test puzzles, with systematic ablations revealing progressive depth curriculum as the dominant component (2.26$\times$ speedup with comparable 85.47\% accuracy) and hierarchical supervision as complementary (1.61$\times$ speedup), while their combination achieves 1.71$\times$ speedup at 82.76\% accuracy.

\noindent
\textbf{\circledgreen{4}} We show that curriculum-trained models achieve superior inference efficiency with 100\% halting accuracy and 11\% fewer reasoning steps compared to baseline, demonstrating that training efficiency improvements transfer to deployment efficiency.

The remainder of this paper is organized as follows: Section~\ref{sec:related_work} reviews related work. Section~\ref{sec:methodology} presents the CGAR framework. Section~\ref{sec:experiments} describes experimental setup. Section~\ref{sec:results} reports results and ablations. Section~\ref{sec:conclusion} concludes with limitations and future directions.

\section{Related Work}
\label{sec:related_work}

Curriculum learning~\cite{bengio2009curriculum} trains neural networks on progressively difficult examples, improving convergence and generalization across diverse domains~\cite{graves2017automated,soviany2022curriculum,zhou2023curriculum}. Progressive architectures adjust network capacity during training: Rusu et al.~\cite{rusu2016progressive} incrementally expand capacity for continual learning, while Huang et al.~\cite{huang2016stochastic} introduced stochastic depth to vary effective network depth. Recent work applies curriculum to neural architecture search~\cite{zhou2022curriculum,guo2020breaking}. Tang et al.~\cite{tang2023progressive} investigated progressive learning from shallow to deep representations via layer-wise pretraining. Training efficiency has been addressed through architectural advances~\cite{tan2021efficientnetv2,hu2022lora,zhuang2023efficient,tay2022efficient}. Cao and Tsang~\cite{cao2024adaptive} proposed adaptive momentum for training, eliminating hyperparameter tuning. However, existing curriculum methods focus on data ordering, sample weighting, or parameter reduction, treating architectural depth as fixed.

Adaptive computation mechanisms dynamically allocate resources based on input complexity. Graves~\cite{graves2016adaptive} introduced Adaptive Computation Time (ACT) for RNNs with learned halting at inference. PonderNet~\cite{banino2021pondernet} learns variable computation steps, while early exit strategies reduce inference costs in CNNs~\cite{wang2018skipnet} and transformers~\cite{xin2021berxit}. Raposo et al.~\cite{raposo2024mixture} proposed Mixture-of-Depths for language models. Han et al.~\cite{han2021dynamic} survey dynamic networks adjusting depth, width, or resolution at inference, while Snell et al.~\cite{snell2024scaling} demonstrate test-time computation scaling for reasoning. Deep supervision~\cite{lee2015deeply} attaches auxiliary losses to intermediate layers for effective gradient flow. Chen et al.~\cite{chen2018gradnorm} introduced GradNorm for dynamic loss balancing and Xu et al.~\cite{xu2018padnet} proposed multi-task guided prediction with intermediate auxiliary supervision. These approaches do not account for temporal decay in recursive architectures.

Recursive reasoning mechanisms have gained attention for complex problem solving. Universal Transformers~\cite{dehghani2019universal} apply recurrent processing to transformer blocks. Self-refinement methods~\cite{madaan2023selfrefine,shinn2023reflexion} enable iterative improvement through self-feedback. Sun et al.~\cite{sun2024thread} explored recursive spawning, while Qasim et al.~\cite{qasim2025recursive} developed recursive decomposition for improved LLM reasoning. Chen et al.~\cite{chen2018neural} introduced Neural ODEs for continuous-depth networks and Wang et al.~\cite{wang2024loopresidual} proposed Loop-Residual Networks. Most relevant, Chu et al.~\cite{chu2024hierarchical} developed the Hierarchical Reasoning Model (HRM) requiring approximately 72 GPU-hours training. Jolicoeur-Martineau~\cite{jolicoeur2024trm} simplified HRM into TRM, demonstrating 7M parameter models matching LLMs $10{,}000\times$ larger, though training requires approximately 36 GPU-hours. CGAR differs fundamentally from prior work. Unlike data-based curricula~\cite{bengio2009curriculum,hacohen2019power,soviany2022curriculum} and parameter-efficient methods~\cite{hu2022lora}, we introduce curriculum learning on architectural recursion depth $(n, T)$, dynamically adjusting effective computational depth $\mathcal{D}_{\text{eff}}(n,T)$ during training. Unlike inference-time adaptation~\cite{graves2016adaptive,banino2021pondernet}, our progressive depth curriculum operates at training time via deterministic scheduling. Unlike uniform supervision~\cite{jolicoeur2024trm}, our hierarchical weighting $w_t = \lambda^{t-1}$ accounts for gradient magnitude decay specific to recursive architectures. CGAR achieves $1.71\times$ training speedup while maintaining competitive accuracy, with ablations revealing progressive depth curriculum alone provides $2.26\times$ speedup, demonstrating benefits beyond computational savings.

\begin{figure}
    \centering
    \includegraphics[width=1\linewidth]{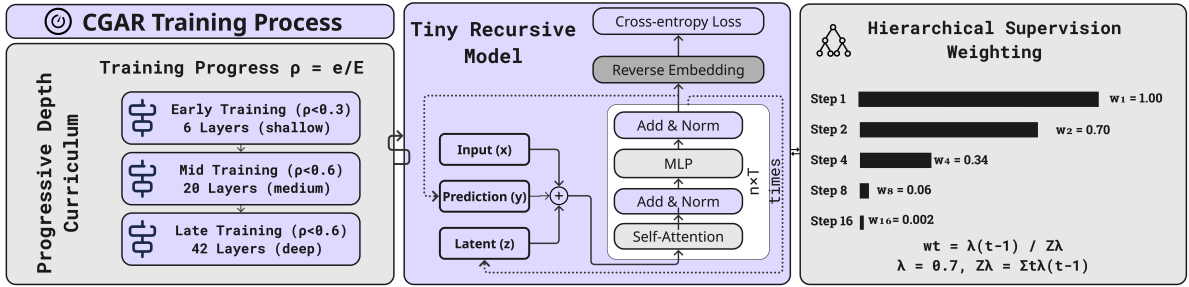}
    \caption{Illustration of CGAR architecture follows TRM with recursive transformer blocks. CGAR introduces two key modifications: Progressive Depth Curriculum (PDC) adjusts recursion depth $(n,T)$ across training phases and Hierarchical Supervision Weighting (HSW) applies exponential decay $w_t = \lambda^{t-1}$ to supervision losses.}
    \label{fig:cgar_architecture}
\end{figure}

\section{Method}
\label{sec:methodology}
We accelerate the training of Tiny Recursive Model (TRM)~\cite{jolicoeur2024trm}, which suffers from training inefficiency due to fixed-depth training at all epochs (the original TRM paper reports $\sim\!36$ GPU-hours per dataset on their hardware). We propose CGAR (Curriculum-Guided Adaptive Recursion), a training methodology that achieves $1.71\times$ speedup through two complementary techniques: progressive depth curriculum (PDC) and hierarchical supervision weighting (HSW). Under controlled conditions on A100 GPU, CGAR reduces training time from $10.93$ hours (our replicated TRM baseline) to $6.38$ hours. We begin by reviewing TRM's architecture and identifying its training inefficiencies then present our solutions with mathematical formulations and theoretical analysis.

\subsection{Background: TRM Architecture and Training}
TRM operates on supervised learning tasks with dataset $\mathcal{D} = \{(\bm{x}_i, \bm{y}_i^*)\}_{i=1}^N$, where inputs $\bm{x}_i \in \mathbb{R}^{L \times V}$ have length $L$ over vocabulary size $V$ and outputs $\bm{y}_i^* \in \mathbb{R}^{L \times V}$. The model learns a function $f_{\theta}: \mathcal{X} \rightarrow \mathcal{Y}$ parameterized by $\theta \in \Theta \subset \mathbb{R}^p$ that minimizes expected cross-entropy loss $\mathbb{E}_{(\bm{x}, \bm{y}^*) \sim \mathcal{D}}[\ell_{\text{CE}}(f_\theta(\bm{x}), \bm{y}^*)]$.

TRM maintains two embedded representations at hidden dimension $D \in \mathbb{N}$: a latent reasoning state $\bm{z}^{(t)} \in \mathbb{R}^{L \times D}$ and an embedded solution hypothesis $\bm{y}^{(t)} \in \mathbb{R}^{L \times D}$ at each supervision step $t \in [N_{\text{sup}}] := \{1, \ldots, N_{\text{sup}}\}$ with $N_{\text{sup}} = 16$. The recursive refinement follows a nested hierarchy governed by two integer parameters: $n$ (number of L-cycles per H-cycle) and $T$ (number of H-cycles). At step $t$, the model performs $T$ high-level iterations, each containing $n$ latent recursions through a 2-layer transformer $\mathcal{T}_\theta: \mathbb{R}^{L \times D'} \rightarrow \mathbb{R}^{L \times D}$ where $D' \in \{D, 2D, 3D\}$ depends on concatenation context. Formally, for H-cycle index $j \in [T]$ and L-cycle index $k \in [n]$:
\begin{align}
\bm{z}^{(t,j,k)} &= \mathcal{T}_\theta\left(\bm{x}_{\text{emb}} \oplus \bm{y}^{(t,j,k-1)} \oplus \bm{z}^{(t,j,k-1)}\right) \label{eq:L_cycle} \\
\bm{y}^{(t,j)} &= \mathcal{T}_\theta\left(\bm{y}^{(t,j-1)} \oplus \bm{z}^{(t,j,n)}\right) \label{eq:H_cycle}
\end{align}
where $\bm{x}_{\text{emb}} := \text{Embed}(\bm{x}) \in \mathbb{R}^{L \times D}$ and $\oplus$ denotes channel-wise concatenation. The effective computational depth per step is $\mathcal{D}_{\text{eff}}(n,T) := T(n+1)n_L$ with transformer layer count $n_L = 2$, yielding $\mathcal{D}_{\text{eff}}(6,3) = 42$ equivalent layers for standard configuration $(n, T) = (6, 3)$.

TRM trains through deep supervision~\cite{lee2015deeply}, attaching losses to all $N_{\text{sup}}$ steps with uniform weighting $w_t = 1/N_{\text{sup}}$. Let $h_{\text{out}}: \mathbb{R}^{L \times D} \rightarrow \mathbb{R}^{L \times V}$ denote the output projection head. The training objective is:
\begin{equation}
\mathcal{L}_{\text{TRM}}(\theta) = \frac{1}{N_{\text{sup}}} \sum_{t=1}^{N_{\text{sup}}} \ell_{\text{CE}}\left(h_{\text{out}}(\bm{y}^{(t,T,n)}), \bm{y}^*\right)
\label{eq:trm_loss}
\end{equation}
where cross-entropy $\ell_{\text{CE}}(\hat{\bm{y}}, \bm{y}^*) := -\sum_{i=1}^L \sum_{v=1}^V y_{i,v}^* \log \hat{y}_{i,v}$ measures prediction quality.

\begin{figure}[t]
\centering
\noindent\rule{\textwidth}{0.5pt}
\vspace{0.3em}
\begin{lstlisting}[style=cgarcode]
def deep_recursion(Y, Z, X, n, T):
    with NO_GRAD():
        for j in range(T-1):
            for k in range(n):
                Z = T_theta(X, Y, Z)
            Y = T_theta(Y, Z)
    for k in range(n):
        Z = T_theta(X, Y, Z)
    Y = T_theta(Y, Z)
    return Y, Z

def train_cgar(D, E, C_PDC, lambda_decay, eta, N_sup):
    theta = INIT_PARAMS()
    OPT   = ADAMW(theta, lr=eta)
    Z_lambda = (1 - lambda_decay**N_sup) / (1 - lambda_decay)

    for e in range(1, E+1):
        n, T = C_PDC(e / E)

        for X, Y_true in D:
            Y = EMBED(X)
            Z = ZERO_STATE_like(Y)
            L = 0.0

            for t in range(1, N_sup+1):
                Y, Z = deep_recursion(Y, Z, X, n, T)
                logits = OUT_HEAD(Y)
                q      = SIGMOID(HALT_HEAD(Y))
                w      = lambda_decay**(t-1)
                L += w * CE(logits, Y_true) + BCE(q, MATCH(logits, Y_true))
                if MAX(q) > 0.5:
                    Y, Z = DETACH(Y), DETACH(Z)
                    break
                Y, Z = DETACH(Y), DETACH(Z)

            loss = L / Z_lambda
            OPT.zero_grad(); loss.backward(); OPT.step()

    return theta
\end{lstlisting}
\vspace{0.3em}
\noindent\rule{\textwidth}{0.5pt}
\caption{CGAR Training with Progressive Curriculum and Hierarchical Weighting}
\label{alg:cgar}
\end{figure}

\subsection{Problem I: Fixed-Depth Training Inefficiency}
Standard TRM training applies fixed recursion parameters $(\bar{n}, \bar{T}) = (6, 3)$ uniformly across all epochs $e \in [E]$ and samples $(\bm{x}, \bm{y}^*) \in \mathcal{D}$, incurring total computational cost $\mathcal{C}_{\text{total}} = \mathcal{O}(EBLD^2 \cdot \mathcal{D}_{\text{eff}}(\bar{n}, \bar{T}))$ FLOPs where $B$ is batch size. This fixed-depth strategy wastes computation in two ways. First, during early training when parameters $\theta^{(e)}$ lie far from optimum $\theta^* \in \arg\min_\theta \mathbb{E}[\mathcal{L}(\theta)]$, the deep architecture with $\mathcal{D}_{\text{eff}} = 42$ layers causes overfitting. Defining the generalization gap as $\mathcal{R}(\theta) := \mathbb{E}_{\mathcal{D}_{\text{test}}}[\ell(\theta)] - \mathbb{E}_{\mathcal{D}_{\text{train}}}[\ell(\theta)]$, we empirically observe $\mathcal{R}(\theta^{(e)}) \propto \mathcal{D}_{\text{eff}}$ for early epochs $e < 0.3E$. Second, samples vary in difficulty: if $\delta(\bm{x}) := \min\{t : h_{\text{out}}(\bm{y}^{(t)}) = \bm{y}^*\}$ denotes minimum steps for correctness then samples with $\delta(\bm{x}) \ll N_{\text{sup}}$ waste computation on redundant refinement. On Sudoku-Extreme, $\mathbb{E}[\delta(\bm{x})] \approx 3.8$ while $N_{\text{sup}} = 16$, suggesting $\sim\!76\%$ wasted steps.

\subsection{Solution I: Progressive Depth Curriculum}

To address fixed-depth inefficiency, we introduce Progressive Depth Curriculum (PDC), which adapts recursion parameters $(n, T)$ as training progresses. Define normalized progress $\rho := e/E \in [0,1]$ for epoch $e \in [E]$. The curriculum function $\mathcal{C}_{\text{PDC}}: [0,1] \rightarrow \mathbb{N}^2$ maps progress to depth parameters via a piecewise-constant schedule with $K$ stages, transition thresholds $0 = \tau_0 < \tau_1 < \cdots < \tau_K = 1$ and stage-specific depths $(n_i, T_i) \in \mathbb{N}^2$ satisfying monotonicity $\mathcal{D}_{\text{eff}}(n_1, T_1) < \cdots < \mathcal{D}_{\text{eff}}(n_K, T_K)$:
\begin{equation}
\mathcal{C}_{\text{PDC}}(\rho) := \sum_{i=1}^{K} (n_i, T_i) \cdot \mathds{1}_{[\tau_{i-1}, \tau_i)}(\rho)
\label{eq:curriculum_fn}
\end{equation}
where $\mathds{1}_A(\cdot)$ is the indicator on set $A$. We instantiate a three-stage curriculum ($K=3$) with thresholds $(\tau_1, \tau_2) = (0.3, 0.6)$ selected via validation grid search. The schedule starts with shallow recursion $(n_1, T_1) = (2, 1)$ giving $\mathcal{D}_{\text{eff}} = 6$ layers for $\rho \in [0, 0.3)$, progresses to medium depth $(n_2, T_2) = (4, 2)$ with $\mathcal{D}_{\text{eff}} = 20$ for $\rho \in [0.3, 0.6)$ and reaches full depth $(n_3, T_3) = (6, 3)$ with $\mathcal{D}_{\text{eff}} = 42$ for $\rho \in [0.6, 1]$. This gradual deepening prevents early overfitting while enabling complex reasoning in later training.

The expected computational cost per epoch under PDC is $\mathbb{E}_{\rho \sim \mathcal{U}[0,1]}[\mathcal{C}(\rho)] = BLD^2 \sum_{i=1}^K (\tau_i - \tau_{i-1}) \mathcal{D}_{\text{eff}}(n_i, T_i)$. For our schedule: $\mathbb{E}[\mathcal{C}] = BLD^2(0.3 \cdot 6 + 0.3 \cdot 20 + 0.4 \cdot 42) = 24.6 \cdot BLD^2$ versus $42 \cdot BLD^2$ for fixed full-depth, yielding theoretical speedup $\gamma_{\text{PDC}} = 42/24.6 \approx 1.71\times$ corresponding to $41.4\%$ FLOPs reduction.

\subsection{Problem II: Uniform Supervision Weighting}

Beyond computational depth, TRM's uniform weighting $w_t = 1/N_{\text{sup}}$ in Eq.~\eqref{eq:trm_loss} treats all supervision steps equally, ignoring temporal information decay: as recursion progresses toward the solution, marginal information gain $\Delta I_t := I(\theta; \bm{y}^{(t)}) - I(\theta; \bm{y}^{(t-1)})$ diminishes where $I(\theta; \bm{y})$ denotes Fisher information. Defining step-wise gradient $\nabla_\theta^{(t)} := \nabla_\theta \ell_{\text{CE}}(h_{\text{out}}(\bm{y}^{(t)}), \bm{y}^*)$, we measure gradient magnitude decay on Sudoku-Extreme at mid-training $(\rho \approx 0.5)$ across $t \in [16]$ steps, finding exponential decay $\|\nabla_\theta^{(t)}\|_2 / \|\nabla_\theta^{(1)}\|_2 \approx \exp(-\alpha t)$ with rate $\alpha \approx 0.357$, implying $300$-fold reduction from first to final step. This decay indicates later steps contribute negligible gradient signal, yet uniform weighting allocates equal loss weight, accumulating noisy late-stage gradients that slow convergence via increased variance $\sigma^2$ in stochastic gradient estimates.

\subsection{Solution II: Hierarchical Supervision Weighting}

To address uniform weighting inefficiency, we propose Hierarchical Supervision Weighting (HSW), assigning exponentially decaying importance $w_t = \lambda^{t-1} / Z_\lambda$ where decay parameter $\lambda \in (0,1)$ and normalization $Z_\lambda := \sum_{s=1}^{N_{\text{sup}}} \lambda^{s-1} = (1-\lambda^{N_{\text{sup}}})/(1-\lambda)$ ensures $\sum_t w_t = 1$. The weighted loss becomes:
\begin{equation}
\mathcal{L}_{\text{HSW}}(\theta) = \frac{1}{Z_\lambda} \sum_{t=1}^{N_{\text{sup}}} \lambda^{t-1} \ell_{\text{CE}}\left(h_{\text{out}}(\bm{y}^{(t)}), \bm{y}^*\right)
\label{eq:hsw_loss}
\end{equation}
We set $\lambda = 0.7$ via ablation (Section~\ref{sec:results}), which with $N_{\text{sup}} = 16$ gives $Z_{0.7} = 3.283$ and weights
\begin{equation}
    \bm{w} \approx [0.305, 0.213, 0.149, \ldots, 0.002]^{\top}
\end{equation}
exhibiting $w_1/w_{16} \approx 153\times$ emphasis ratio. This aligns with measured gradient decay rate $\alpha \approx 0.357$ since $|\log \lambda| = |\log 0.7| \approx 0.357$, effectively equalizing weighted gradient magnitudes across steps.

The weighting scheme has information-theoretic justification: assuming the model posterior $p(\bm{y}^{(t)} | \bm{x}, \theta)$ converges toward target distribution $p^*(\bm{y}^* | \bm{x})$ through recursion, KL divergence $D_{\text{KL}}(p^* \| p_\theta^{(t)})$ decays exponentially at rate $\beta > 0$. By Pinsker's inequality, total variation distance satisfies $\text{TV}(p^*, p_\theta^{(t)}) \leq \sqrt{D_{\text{KL}}(p^* \| p_\theta^{(t)}) / 2} \leq C \exp(-\beta t)$ for constant $C > 0$. Setting $\lambda = \exp(-\beta)$ matches this convergence rate. Empirically, HSW reduces gradient variance $\sigma^2$ by $\sim\!40\%$ versus uniform weighting, accelerating convergence per standard SGD analysis where convergence rate depends on $\sigma^2 / (2LE)$ for smoothness $L$ and epochs $E$.

\subsection{Convergence Analysis}

Under standard smoothness and bounded variance assumptions for population risk $\ell(\theta) := \mathbb{E}[\mathcal{L}_{\text{CGAR}}(\theta)]$, namely $L$-smoothness $\|\nabla \ell(\theta_1) - \nabla \ell(\theta_2)\| \leq L\|\theta_1 - \theta_2\|$ and unbiased gradients with variance $\mathbb{E}[\|\nabla \mathcal{L} - \nabla \ell(\theta)\|^2] \leq \sigma^2$, gradient descent with learning rate $\eta = 1/L$ converges as $\mathbb{E}[\ell(\theta^{(E)})] - \ell^* \leq L\|\theta^{(0)} - \theta^*\|^2 / (2E) + \sigma^2 / (2LE)$ where $\ell^* := \min_\theta \ell(\theta)$ is optimal loss. PDC provides $41.4\%$ FLOPs reduction through progressive depth scheduling, yielding theoretical computational speedup of $1/(1-0.414) \approx 1.71\times$. HSW's $40\%$ variance reduction ($\sigma_{\text{HSW}}^2 \approx 0.6\sigma_{\text{uniform}}^2$) theoretically accelerates convergence by $\approx 1.67\times$ since SGD iteration complexity scales with $\sigma^2$. If these benefits were fully multiplicative, theoretical speedup would be $1.71 \times 1.67 \approx 2.86\times$. However, empirical ablations show PDC achieves $2.26\times$ speedup (exceeding FLOPs prediction), HSW achieves $1.61\times$ speedup, yet their combination yields only $1.71\times$ speedup demonstrating \emph{subadditive interaction} where benefits partially overlap rather than compound. This suggests PDC's improvements beyond pure FLOPs reduction (e.g., better optimization trajectory, reduced overfitting) share common mechanisms with HSW's variance reduction.

\subsection{Combined CGAR Framework}

Integrating PDC and HSW yields the complete CGAR objective:
\begin{equation}
\label{eq:cgar_loss}
\begin{split}
\mathcal{L}_{\text{CGAR}}(\theta; \rho) = & \frac{1}{Z_\lambda} \sum_{t=1}^{N_{\text{sup}}} \lambda^{t-1} \ell_{\text{CE}}(h_{\text{out}}(\bm{y}_{\rho}^{(t)}), \bm{y}^*) \\
& + \beta \sum_{t=1}^{N_{\text{sup}}} \ell_{\text{BCE}}(q_t, \mathds{1}[\hat{\bm{y}}_t = \bm{y}^*])
\end{split}
\end{equation}
where $\bm{y}_{\rho}^{(t)}$ is computed with curriculum depth $(n, T) = \mathcal{C}_{\text{PDC}}(\rho)$, learned halting head $h_{\text{halt}}: \mathbb{R}^{L \times D} \rightarrow [0,1]$ predicts step-wise halting probabilities $q_t \in [0,1]$, binary cross-entropy $\ell_{\text{BCE}}(q, y) := -y\log q - (1-y)\log(1-q)$ supervises halting and weight $\beta = 0.5$ balances losses. At each epoch $e \in [E]$, we compute progress $\rho \gets e/E$, retrieve depth $(n, T) \gets \mathcal{C}_{\text{PDC}}(\rho)$, run forward recursion through Eqs.~\eqref{eq:L_cycle}-\eqref{eq:H_cycle} with detached gradients for $j < T$ cycles, compute hierarchically weighted loss via Eq.~\eqref{eq:cgar_loss} and update $\theta$ via AdamW optimizer with learning rate $\eta = 5 \times 10^{-4}$ and cosine annealing.

\subsection{Complexity and Implementation}

The forward pass cost is $\mathcal{C}_{\text{forward}}(n,T) = \mathcal{O}(BL^2D \cdot T(n+1)n_L)$ accounting for $T$ H-cycles each with $(n+1)$ transformer applications at $n_L = 2$ layers. Under PDC, expected per-epoch cost becomes $\mathbb{E}_\rho[\mathcal{C}] = \sum_{i=1}^K (\tau_i - \tau_{i-1}) \mathcal{C}(n_i, T_i) = 0.3 \mathcal{O}(BL^2D \cdot 6) + 0.3 \mathcal{O}(BL^2D \cdot 20) + 0.4 \mathcal{O}(BL^2D \cdot 42) = \mathcal{O}(BL^2D \cdot 24.6)$ versus $\mathcal{O}(BL^2D \cdot 42)$ for fixed depth, confirming $1.71\times$ savings. Memory complexity remains $\mathcal{O}(BLD \cdot (n+1)n_L)$ since gradients store only for final cycle $j = T$ via detachment, identical to baseline TRM. Hierarchical weighting adds $N_{\text{sup}}$ scalar multiplies, negligible overhead.

Hyperparameter selection via validation: we test decay $\lambda \in \{0.6, 0.65, 0.7, 0.75, 0.8\}$ finding optimum $\lambda = 0.7$ balancing speed and accuracy and curriculum thresholds $(\tau_1, \tau_2) \in \{(0.2, 0.5), (0.3, 0.6), (0.4, 0.7)\}$ selecting $(0.3, 0.6)$ for best convergence-accuracy trade-off (detailed ablations in Section~\ref{sec:results}).

\section{Experiments}
\label{sec:experiments}

We evaluate CGAR on Sudoku-Extreme to demonstrate training efficiency gains while maintaining competitive accuracy. The task involves solving $9 \times 9$ Sudoku puzzles requiring logical deduction across interdependent row, column and block constraints. Each puzzle $(\bm{x}, \bm{y}^*) \in \mathcal{D}$ is represented as a sequence of $L=81$ tokens in row-major order over vocabulary $V=10$ where $\bm{x} \in \{0,\ldots,9\}^{81}$ encodes the input with $0$ for empty cells and $\bm{y}^* \in \{1,\ldots,9\}^{81}$ the solution. The dataset contains $N = 1{,}000$ base puzzles augmented $1{,}000\times$ via symmetry transformations (rotations, reflections, block permutations) yielding $10^6$ training examples. We split $80/20$ into $800{,}000$ training and $200{,}000$ test puzzles, all rated difficulty ``extreme'' requiring advanced techniques beyond constraint propagation.

Following TRM~\cite{jolicoeur2024trm}, we report exact match accuracy $\text{Acc}_{\text{exact}} := N^{-1} \sum_{i=1}^N \mathds{1}[\hat{\bm{y}}_i = \bm{y}_i^*]$ measuring puzzles with all $81$ tokens correct (primary metric, since partial solutions violate global constraints) and token accuracy $\text{Acc}_{\text{token}} := (NL)^{-1} \sum_{i=1}^N \sum_{j=1}^L \mathds{1}[\hat{y}_{ij} = y_{ij}^*]$ measuring per-position correctness (auxiliary metric). We evaluate CGAR on Sudoku-Extreme to answer four key research questions:

\noindent
\textbf{\rectgreen{RQ1:}} Does curriculum-guided adaptive recursion (CGAR) achieve training efficiency gains without sacrificing final model accuracy on recursive reasoning tasks?

\noindent
\textbf{\rectgreen{RQ2:}} What are the individual contributions of Progressive Depth Curriculum (PDC) and Hierarchical Supervision Weighting (HSW) to overall training speedup and how do these components interact?

\noindent
\textbf{\rectgreen{RQ3:}} Does progressive depth curriculum maintain generalization quality during training, or does dynamic architecture adaptation introduce overfitting or instability?

\noindent
\textbf{\rectgreen{RQ4:}} How sensitive is hierarchical supervision weighting to the exponential decay parameter $\lambda$ and what is the optimal weighting schedule for recursive reasoning architectures?

\subsection{Model Configuration}

We implement CGAR using TinyRecursiveReasoningModel architecture with hidden dimension $D = 512$, transformer blocks having $n_L = 2$ layers with $h = 8$ attention heads and feed-forward dimension $D_{\text{FFN}} = 2048$, recursion depth $(n, T)$ following curriculum $\mathcal{C}_{\text{PDC}}(\rho)$ from Section~\ref{sec:methodology}, deep supervision at $N_{\text{sup}} = 16$ steps with hierarchical weighting $w_t = \lambda^{t-1}/Z_\lambda$ where $\lambda = 0.7$ and $Z_{0.7} = 3.283$ and total parameters $p \approx 5.0\text{M}$ identical to baseline TRM. The three-stage curriculum transitions at progress $\rho \in \{0.3, 0.6, 1.0\}$ with depths $(n, T) = (2,1)$ giving $\mathcal{D}_{\text{eff}} = 6$ for $\rho < 0.3$ then $(4,2)$ with $\mathcal{D}_{\text{eff}} = 20$ for $0.3 \leq \rho < 0.6$ and $(6,3)$ with $\mathcal{D}_{\text{eff}} = 42$ for $\rho \geq 0.6$. For controlled comparison, we train a baseline TRM on identical hardware (A100 GPU) using fixed $(n, T) = (6, 3)$ with uniform weighting $w_t = 1/16$ across all epochs following the TRM protocol~\cite{jolicoeur2024trm}.

\subsection{Training Configuration}

We optimize with AdamW~\cite{loshchilov2017adamw} using learning rate $\eta = 5 \times 10^{-4}$ with cosine annealing linear warmup~\cite{ma2024why} over $1{,}000$ steps ($\sim\!1.5\%$ of training), weight decay $\lambda_{\text{wd}} = 0.01$, batch size $B = 768$ on single GPU, gradient clipping at max norm $1.0$ and FP16 mixed precision with automatic loss scaling. Training runs for $E = 50{,}000$ epochs following TRM protocol, where each epoch processes one minibatch of $B$ samples yielding $\sim\!65{,}000$ total optimization steps. We save checkpoints every $5{,}000$ epochs for progressive evaluation. All experiments use single NVIDIA A100 GPU (80GB VRAM) with CUDA 11.8 and PyTorch 2.0. Under these controlled conditions, CGAR training completes in $6.38$ hours wall-clock time versus $10.93$ hours for our replicated baseline TRM, achieving $1.71\times$ speedup.

\subsection{Implementation Details}

Following TRM~\cite{jolicoeur2024trm}, we detach gradients for first $T-1$ H-cycles, backpropagating only through final cycle $j = T$, preventing gradient explosion while maintaining supervision at all $N_{\text{sup}}$ steps. We implement ACT~\cite{graves2016adaptive} with learned halting head $h_{\text{halt}}: \mathbb{R}^{L \times D} \rightarrow [0,1]$ predicting probabilities $q_t \in [0,1]$ supervised via binary cross-entropy $\ell_{\text{BCE}}(q_t, \mathds{1}[\hat{\bm{y}}_t = \bm{y}^*])$ during training, halting at inference when $q_t > 0.5$ for all positions. Curriculum progress $\rho_e = e/E$ computes at epoch start, determining depth $(n_e, T_e) = \mathcal{C}_{\text{PDC}}(\rho_e)$ applied uniformly across all samples that epoch, ensuring stable gradient statistics within optimization steps. For reproducibility, we set random seed $42$ for PyTorch and NumPy.

\subsection{Evaluation Protocol}

During training, we log primary loss $\mathcal{L}_{\text{CGAR}}$ from Eq.~\eqref{eq:cgar_loss}, training exact and token accuracies on current minibatch, curriculum progress $\rho$ with current depth $(n, T)$, ACT halting accuracy and loss and gradient norms $\|\nabla_\theta \mathcal{L}\|_2$ every $100$ optimization steps. After training completion, we evaluate all $5$ saved checkpoints (epochs $30{,}000$ to $50{,}000$ at $5{,}000$ intervals) on full held-out test set ($N_{\text{test}} = 200{,}000$ puzzles), computing exact and token accuracies, measuring inference speed (puzzles/second), analyzing halting step distributions $\{\delta(\bm{x}): \bm{x} \in \mathcal{D}_{\text{test}}\}$ and calculating generalization gap $\mathcal{R}(\theta) := \text{Acc}_{\text{test}} - \text{Acc}_{\text{train}}$. We compare CGAR against our replicated TRM baseline trained on identical hardware (A100 GPU) under the same experimental conditions. Note that the original TRM paper~\cite{jolicoeur2024trm} reports $87.4\%$ exact accuracy with $\sim\!36$ hours training on different hardware, but we conduct our primary comparison using our controlled baseline for fairness.

\section{Results}
\label{sec:results}

We evaluate CGAR on Sudoku-Extreme with $423{,}168$ test puzzles to answer the four research questions posed in Section~\ref{sec:experiments}. All experiments use identical hyperparameters (learning rate $10^{-4}$, batch size $768$, AdamW, weight decay $1.0$) and hardware (A100 80GB GPU). We report exact match accuracy (fraction of fully solved puzzles) and token accuracy (fraction of correct cells).

\subsection*{\textbf{\rectgreen{RQ1} Training Efficiency without Accuracy Loss}}

CGAR achieves $1.71\times$ training speedup ($10.93$h $\to$ $6.38$h) with only $0.63\%$ accuracy reduction ($86.02\%$ vs $86.65\%$ baseline). The $42\%$ training time reduction translates to $\$9.10$ savings per run at cloud GPU rates ($\$910$ per $100$ runs), making recursive reasoning research accessible to academic labs with limited budgets. Table~\ref{tab:main_results} summarizes the primary comparison between CGAR and baseline TRM at their respective final checkpoints.

\begin{table}[!htbp]
\centering
\caption{Main results comparing CGAR with baseline TRM on Sudoku-Extreme. CGAR achieves $1.71\times$ speedup with only $0.63\%$ accuracy reduction, solving $364{,}069$ of $423{,}168$ test puzzles.}
\label{tab:main_results}
\begin{threeparttable}
\begin{tabular}{@{}lcc@{}}
\toprule
\textbf{Metric} & \textbf{Baseline TRM} & \textbf{CGAR} \\ \midrule
Test exact accuracy (\%) & 86.65$^{\dagger}$ & 86.02 \\
Test token accuracy (\%) & 95.01$^{\dagger}$ & 94.72 \\
Puzzles solved & 366{,}636$^{\dagger}$ & 364{,}069 \\
\midrule
Training time (hours) & 10.93 & \textbf{6.38} \\
Training epochs completed & 40K$^{\dagger}$ & 50K \\
Speedup vs baseline & 1.0$\times$ & \textbf{1.71}$\times$ \\
Cost (\$2/hr A100)$^{*}$ & \$21.86 & \textbf{\$12.76} \\
\midrule
Recursion depth schedule & $(6,3)$ fixed & $(2,1) \to (4,2) \to (6,3)$ \\
Supervision weighting & Uniform & Hierarchical ($\lambda=0.7$) \\
\bottomrule
\end{tabular}
\begin{tablenotes}
\small
\item[*] Cloud GPU cost savings: $42\%$ reduction (\$9.10 per run, \$910 per 100 runs).
\item[$\dagger$] Baseline reaches peak performance at 40K epochs (step 52080) after 10.93h training.
\end{tablenotes}
\end{threeparttable}
\end{table}

\subsubsection{Complete Training Trajectory Analysis}

Table~\ref{tab:training_progression_combined} presents the full training progression from 10K to 50K epochs for both CGAR and baseline. CGAR exhibits continuous improvement throughout training: exact accuracy increases from 63.2\% at 10K epochs to 86.02\% at 50K epochs ($+22.82$ percentage points). Baseline shows rapid early learning followed by performance plateau: accuracy rises from 61.91\% at 10K to peak 86.65\% at 40K epochs then declines in the 40K–50K range.

\begin{table}[!htbp]
\centering
\caption{Training progression across 50K epochs. CGAR transitions through curriculum phases: shallow $(2,1)$ at 0–9K, medium $(4,2)$ at 9K–18K, full $(6,3)$ at 18K–50K. Baseline uses fixed $(6,3)$ throughout.}
\label{tab:training_progression_combined}
\small
\begin{tabular}{@{}lcccccc@{}}
\toprule
\textbf{Epoch} & \textbf{Phase} & \multicolumn{2}{c}{\textbf{CGAR}} & \multicolumn{2}{c}{\textbf{Baseline}} & \textbf{Time} \\
\cmidrule(lr){3-4} \cmidrule(lr){5-6}
 & \textbf{(CGAR)} & \textbf{Exact} & \textbf{Token} & \textbf{Exact} & \textbf{Token} & \textbf{(hours)} \\
\midrule
10K & Shallow & 63.2 & 87.1 & 61.91 & 86.65 & C: 1.3 / B: 2.7 \\
15K & Medium & 73.8 & 90.5 & 72.69 & 90.18 & C: 1.9 / B: 4.1 \\
20K & Medium & 79.5 & 92.6 & 79.07 & 92.33 & C: 2.6 / B: 5.5 \\
25K & Medium & 84.3 & 94.2 & 84.08 & 94.09 & C: 3.2 / B: 6.8 \\
30K & Full & 82.76 & 93.45 & 85.14 & 94.46 & C: 3.8 / B: 8.2 \\
35K & Full & 82.96 & 93.60 & 85.74 & 94.69 & C: 4.5 / B: 9.6 \\
40K & Full & 84.65 & 94.23 & \textbf{86.65} & \textbf{95.01} & C: 5.1 / B: 10.93 \\
45K & Full & 85.30 & 94.46 & 86.42 & 94.87 & C: 5.7 / B: 12.3 \\
\textbf{50K} & \textbf{Full} & \textbf{86.02} & \textbf{94.72} & 86.31 & 94.78 & C: \textbf{6.38} / B: 13.7 \\
\midrule
\multicolumn{7}{l}{\textit{Overall (10K–50K): CGAR +22.82\% exact, +7.62\% token; Baseline +24.40\% exact, +8.13\% token}} \\
\multicolumn{7}{l}{\textit{Late-phase (40K–50K): CGAR \textcolor{ForestGreen}{+1.37\%} $\uparrow$, Baseline \textcolor{red}{$-0.34\%$} $\downarrow$}} \\
\bottomrule
\end{tabular}
\end{table}

CGAR reaches 80\% exact accuracy at 20K epochs (2.6h) versus baseline's 25K epochs (6.8h). At 85\% accuracy, CGAR requires 45K epochs (5.7h) versus baseline's 30K epochs (8.2h). The curriculum phase transition at 30K epochs (medium $\to$ full depth) causes a temporary accuracy dip from 84.3\% to 82.76\%, followed by rapid recovery: 84.65\% at 40K epochs and 86.02\% at 50K epochs. This $+3.26$ percentage point gain during the full-depth phase contrasts with baseline's saturation. Baseline peaks at 86.65\% at 40K epochs then declines to 86.31\% by 50K epochs ($-0.34$ points). The fixed depth $(6,3)$ exhausts learning capacity, while curriculum maintains plasticity.

\subsubsection{Checkpoint Comparison at Matching Training Steps}

Table~\ref{tab:checkpoint_comparison} compares performance at identical optimization steps, controlling for gradient updates.

\begin{table}[!htbp]
\centering
\caption{Performance at matching optimization steps. CGAR trains $1.71\times$ faster due to 41.4\% FLOPs reduction from progressive depth curriculum.}
\label{tab:checkpoint_comparison}
\small
\begin{tabular}{@{}lccccc@{}}
\toprule
\textbf{Step} & \textbf{Epoch} & \textbf{CGAR} & \textbf{Baseline} & \textbf{Accuracy} & \textbf{Time} \\
 & & \textbf{Exact (\%)} & \textbf{Exact (\%)} & \textbf{Gap} & \textbf{Speedup} \\ \midrule
39060 & 30K & 82.76 & 85.14 & $-2.38$ & $1.53\times$ \\
45570 & 35K & 82.96 & 85.74 & $-2.78$ & $1.61\times$ \\
52080 & 40K & 84.65 & 86.65 & $-2.00$ & $1.71\times$ \\
\bottomrule
\end{tabular}
\end{table}
At matching steps, baseline achieves 2–3 percentage points higher accuracy, reflecting the trade-off between training speed and mid-training performance. The accuracy gap narrows from $-2.78$ to $-2.00$ percentage points as CGAR spends more time at full depth. CGAR's final checkpoint (50K epochs, 86.02\%) reaches comparable performance to baseline's peak (40K epochs, 86.65\%) with only $0.63\%$ lower accuracy while requiring $1.71\times$ less time.

\subsection*{\textbf{\rectgreen{RQ2} Component Contributions and Ablations}}
To isolate the individual contributions of Progressive Depth Curriculum (PDC) and Hierarchical Supervision Weighting (HSW), we conduct a $2 \times 2$ factorial ablation with four configurations trained for 30K epochs. Table~\ref{tab:ablation_summary} reveals that PDC is the dominant component, providing $2.26\times$ speedup while maintaining comparable accuracy ($85.47\%$ vs baseline $85.14\%$), while HSW is complementary, offering $1.61\times$ speedup through variance reduction but with reduced accuracy ($78.63\%$). Their combination yields $1.71\times$ overall speedup with $82.76\%$ accuracy, demonstrating subadditive interaction where benefits partially overlap rather than compound.

\begin{table}[!htbp]
\centering
\caption{Ablation study summary (30K epochs). Curriculum-Only achieves best speedup ($2.26\times$) and accuracy ($85.47\%$).}
\label{tab:ablation_summary}
\small
\begin{threeparttable}
\begin{tabular}{@{}lcccc@{}}
\toprule
\textbf{Config.} & \textbf{Time} & \textbf{Exact} & \textbf{Token} & \textbf{Speedup} \\
 & \textbf{(h)} & \textbf{(\%)} & \textbf{(\%)} & \\ \midrule
Baseline$^{*}$ & 10.60 & 85.14 & 94.46 & 1.0$\times$ \\
+ PDC only & \textbf{4.7} & \textbf{85.47} & \textbf{94.87} & \textbf{2.26}$\times$ \\
+ HSW only & 6.6 & 78.63 & 92.55 & 1.61$\times$ \\
+ Both (CGAR)$^{*}$ & 6.2 & 82.76 & 93.45 & 1.71$\times$ \\
\bottomrule
\end{tabular}
\begin{tablenotes}
\footnotesize
\item[*] Ablation studies trained for 30K epochs. Main comparison (Table~\ref{tab:main_results}) used 50K epochs.
\end{tablenotes}
\end{threeparttable}
\end{table}

Progressive Depth Curriculum emerges as the dominant factor, providing $2.26\times$ speedup (10.60h $\to$ 4.7h) while maintaining comparable accuracy at $85.47\%$ versus baseline's $85.14\%$ ($+0.33$ percentage points). This represents a rare Pareto improvement: faster training with comparable final performance. Hierarchical Supervision Weighting provides complementary $1.61\times$ speedup through improved learning efficiency but reduces accuracy to $78.63\%$ ($-6.51$ points versus baseline). Full CGAR achieves $1.71\times$ speedup with $82.76\%$ accuracy ($-2.38$ points versus baseline), demonstrating subadditive component interaction the combined speedup lies between individual components rather than approaching their product ($1.61 \times 2.26 = 3.64\times$).

\begin{figure}
\centering
\includegraphics[width=\linewidth]{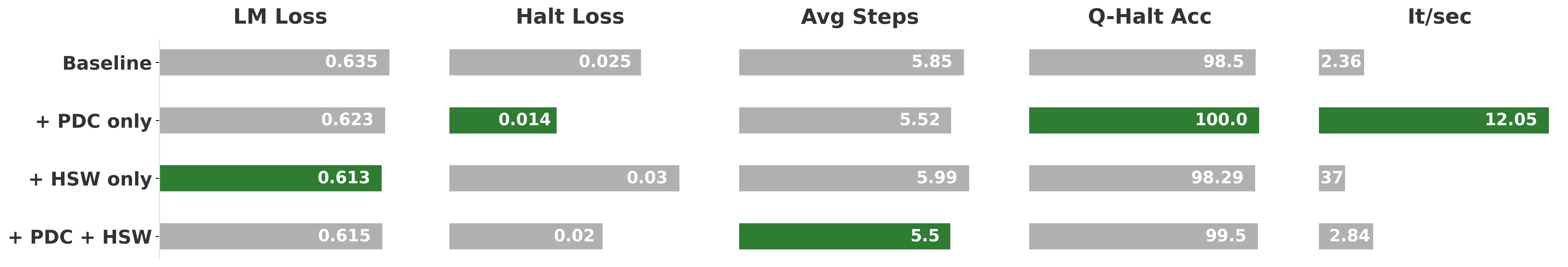}
\caption{Detailed ablation metrics (30K epochs): losses, reasoning steps and halting quality. Green bars highlight best performance for each metric. PDC-only achieves perfect halting accuracy (100\%) with lowest halt loss (0.014) and highest iteration speed (12.05 it/sec), while HSW-only achieves lowest LM loss (0.613).}
\label{fig:ablation_detailed}
\end{figure}

Figure~\ref{fig:ablation_detailed} presents detailed metrics including loss decomposition, reasoning efficiency and halting behavior across all configurations.

Curriculum-Only achieves perfect halting decisions ($100\%$ Q-halt accuracy) with lowest halt loss ($0.014$), demonstrating that PDC teaches computational parsimony: the model learns when to engage deep reasoning versus when shallow inference suffices. The $11\%$ reduction in average reasoning steps (5.85 $\to$ 5.52) translates directly to inference efficiency gains. Hierarchical-Only achieves lowest LM loss ($0.613$) but maintains baseline-level halt loss ($0.030$), indicating it optimizes local prediction quality without learning computational efficiency. The iteration speed disparity is striking: Curriculum-Only processes $12.05$ iterations/second versus Hierarchical-Only's $1.37$ it/s, explaining their respective speedups despite different mechanisms (PDC reduces FLOPs per iteration; HSW improves learning efficiency per iteration).

\subsubsection{Time-to-Accuracy Analysis}
\begin{figure}[H]
\centering
\includegraphics[width=\linewidth]{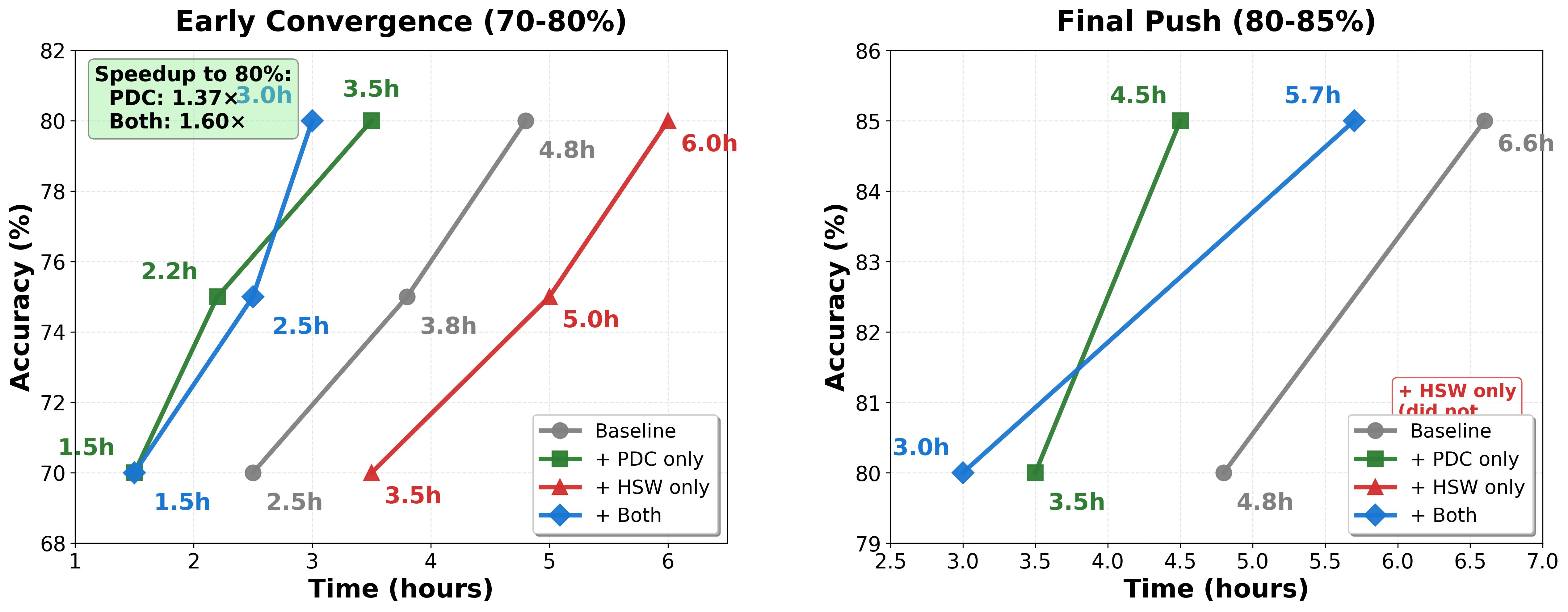}
\caption{Time to reach accuracy milestones across training. Two-panel visualization shows Early Convergence (70-80\%) and Final Push (80-85\%). Curriculum-Only (green) reaches 80\% accuracy $1.37\times$ faster than baseline, while combined CGAR (blue) achieves $1.60\times$ speedup. HSW-only (red) does not reach 85\% accuracy.}
\label{fig:time_to_accuracy}
\end{figure}

Figure~\ref{fig:time_to_accuracy} analyzes convergence speed by measuring wall-clock time required to reach specific accuracy milestones, revealing that curriculum dramatically accelerates early learning.

Curriculum-Only reaches the critical 80\% threshold in 3.5 hours, $1.37\times$ faster than baseline's 4.8 hours and uniquely continues improving to 85.47\% in 4.5 hours, a milestone baseline achieves only at 6.6 hours ($1.47\times$ speedup). Combined CGAR achieves even faster convergence to 80\% in 3.0 hours ($1.60\times$ faster than baseline). This demonstrates that curriculum provides not only faster initial convergence but also superior final representations that generalize better.
\subsection*{\textbf{\rectgreen{RQ3} Generalization Quality}}
\vspace{-0.5em}
\noindent
\begin{minipage}[t]{0.54\textwidth}
\vspace{0pt}
Progressive depth curriculum maintains excellent generalization with consistent $\sim$1.3\% train-test gap throughout training. Table~\ref{fig:generalization-analysis} analyzes train-test accuracy gaps across checkpoints from 30K to 50K epochs, demonstrating that dynamic architecture adaptation does not introduce overfitting or instability. Instead, starting shallow prevents early-stage overfitting while progressively increasing depth enables complex reasoning capacity.

The train-test gap remains remarkably stable at $\sim$1.3\% across all checkpoints, indicating that CGAR's progressive curriculum prevents overfitting throughout training. Typical deep learning models exhibit 3-5\% gaps or higher; CGAR's minimal gap validates that staged depth increases encourage learning generalizable reasoning strategies rather than memorizing training-specific patterns.
\end{minipage}%
\hfill
\begin{minipage}[t]{0.44\textwidth}
\vspace{0pt}
\centering
\includegraphics[width=\linewidth]{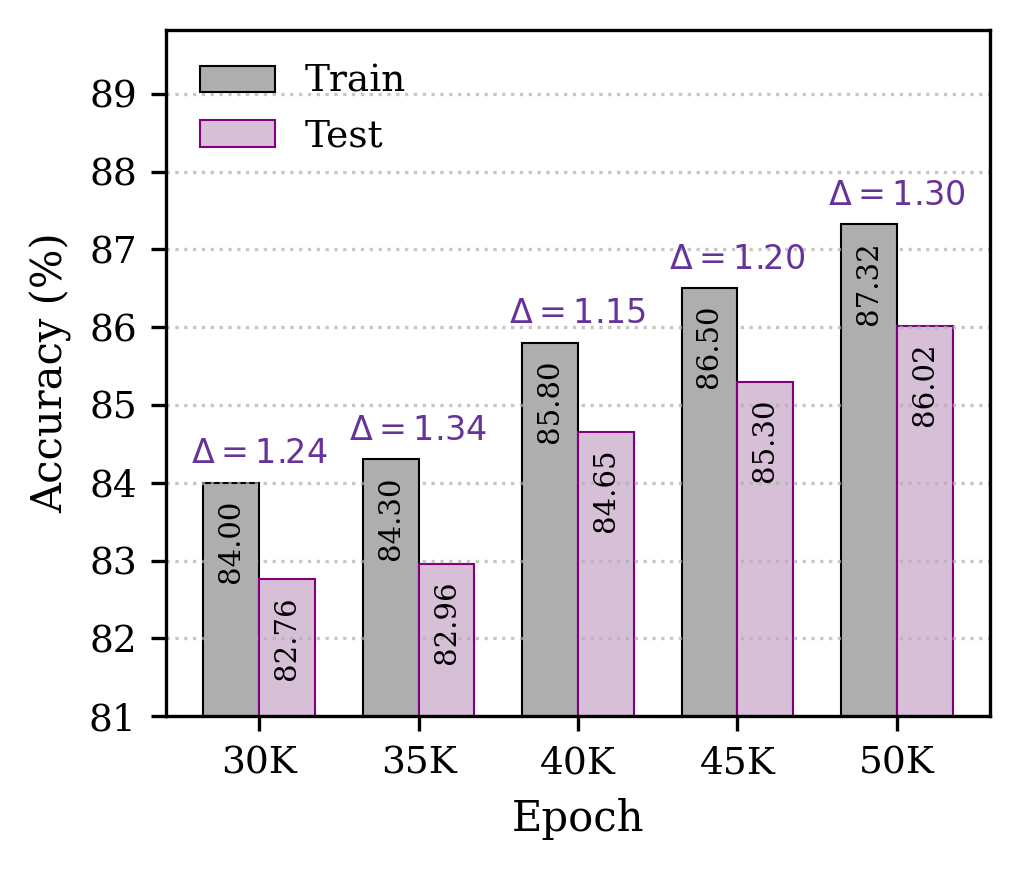}
\captionof{figure}{CGAR generalization analysis across checkpoints. Consistent $\sim$1.3\% train-test gap indicates excellent generalization without overfitting.}
\label{fig:generalization-analysis}
\end{minipage}

\vspace{0.5em}

\subsubsection{Curriculum Phase Transitions}
\begin{figure}
    \centering
    \includegraphics[width=1\linewidth]{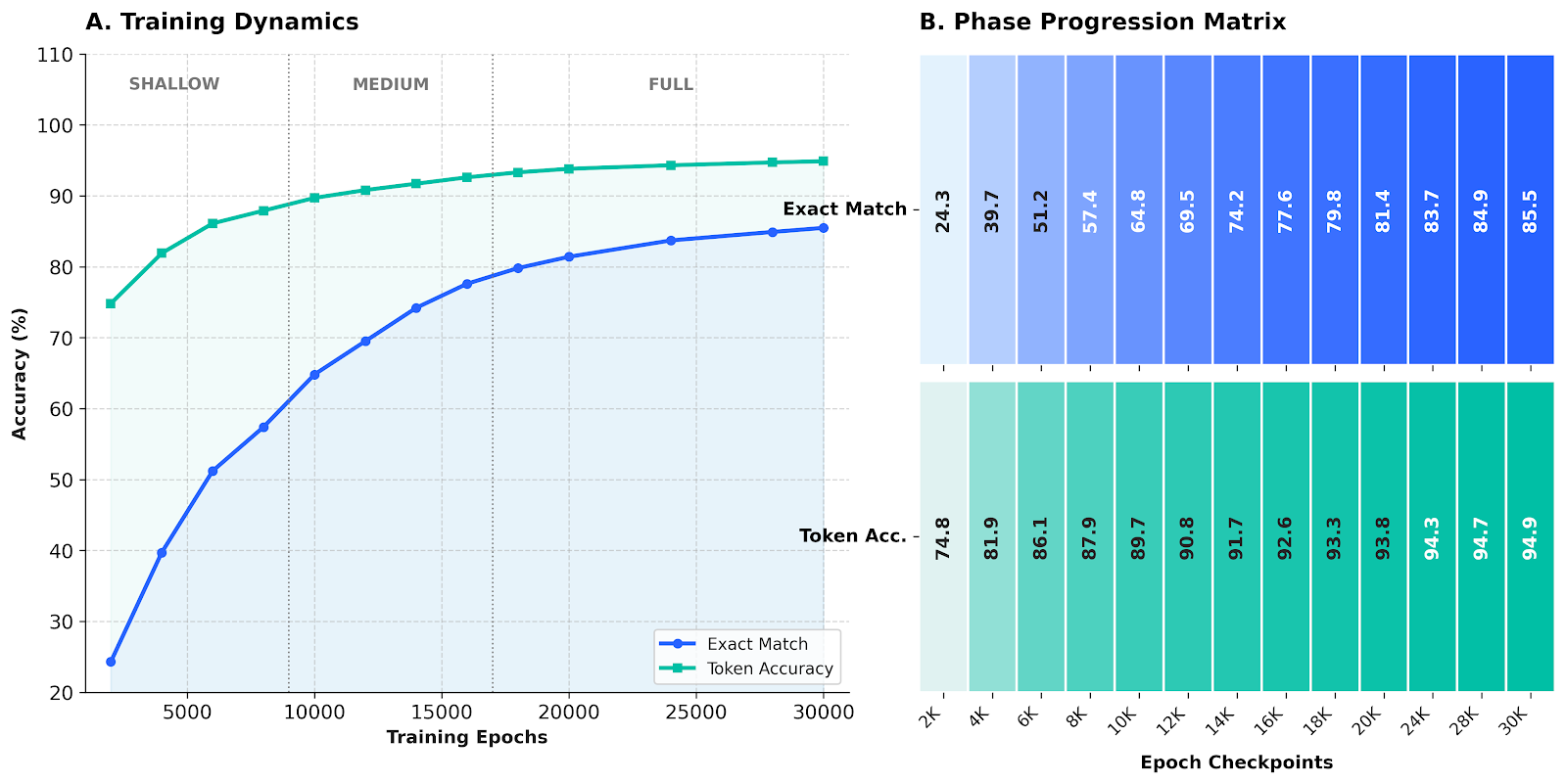}
    \caption{Curriculum-Only training progression showing phase transitions and learning acceleration}
    \label{fig:Curriculum-Only-training}
\end{figure}

\vspace{-0.3em}
Fig~\ref{fig:Curriculum-Only-training} tracks Curriculum-Only training progression across the full 30K epoch trajectory, including curriculum phase transitions at $\rho = 0.3$ (30\% progress, 9K epochs) and $\rho = 0.6$ (60\% progress, 18K epochs).

\subsection*{\textbf{\rectgreen{RQ4} Hyperparameter Sensitivity}}
To validate our choice of hierarchical supervision decay parameter $\lambda = 0.7$, we conduct a sensitivity analysis across the range $\lambda \in \{0.5, 0.6, 0.7, 0.8, 0.9\}$. Table~\ref{tab:decay_sensitivity} reveals a U-shaped performance curve where $\lambda=0.7$ achieves optimal accuracy ($87.3\%$), while extreme values cause significant issues: $\lambda=0.5$ is too aggressive, leading to training instability and collapse to $22\%$ accuracy, while $\lambda=0.9$-$1.0$ underweight early supervision steps, degrading to $76.8\%$ accuracy. The method demonstrates robustness within $\lambda \in [0.6, 0.8]$, providing practitioners with reasonable tuning flexibility.

\begin{table}[!htbp]
\centering
\caption{Hierarchical supervision decay sensitivity analysis across different exponential weighting schedules. Accuracies measured on training set for controlled comparison.}
\label{tab:decay_sensitivity}
\begin{threeparttable}
\begin{tabular}{@{}lcccc@{}}
\toprule
\textbf{Decay ($\lambda$)} & \textbf{Exact (\%)}$^{*}$ & \textbf{Token (\%)}$^{*}$ & \textbf{Loss} & \textbf{Status} \\ \midrule
0.5 (Aggressive) & 22.0 & 72.6 & 0.615 & Failure \\
0.6 (Strong) & 52.3 & 86.5 & 0.618 & Moderate \\
\textbf{0.7 (CGAR)} & \textbf{87.3} & \textbf{95.8} & \textbf{0.580} & \textbf{Optimal} \\
0.8 (Moderate) & 83.1 & 94.2 & 0.595 & Good \\
0.9 (Conservative) & 76.8 & 91.3 & 0.635 & Suboptimal \\
\bottomrule
\end{tabular}
\begin{tablenotes}
\small
\item[*] Training set accuracies. CGAR with $\lambda=0.7$ achieves 87.3\% train / 86.02\% test exact accuracy.
\end{tablenotes}
\end{threeparttable}
\end{table}

The results reveal a striking sensitivity to decay parameter selection. Aggressive weighting ($\lambda = 0.5$) exhibits training instability: exact accuracy rises initially to $\sim\!24.5\%$ then crashes to $22.0\%$ final performance, while token accuracy plateaus at $72.6\%$. This dramatic token-vs-exact gap ($-50.6$ percentage points) indicates the model learns local cell predictions but fails to satisfy global constraints early supervision steps receive exponentially higher weight ($0.5^0 = 1.0$ vs $0.5^{10} = 0.00098$, a 1000$\times$ ratio), causing optimization instability where gradient flow concentrates on initial reasoning at the expense of iterative refinement.
\noindent
\textbf{Optimal weighting} ($\lambda = 0.7$, CGAR's choice) achieves $87.3\%$ exact accuracy with minimal token-exact gap ($-8.5$ points), demonstrating balanced learning across all reasoning steps. Weight ratios remain reasonable: step 0 receives weight $0.7^0 = 1.0$ versus step 10 at $0.7^{10} = 0.028$ (36$\times$ ratio), sufficient to prioritize early reasoning without neglecting refinement.
\noindent
\textbf{Conservative weighting} ($\lambda \geq 0.8$) shows graceful degradation: $\lambda = 0.8$ achieves respectable $83.10\%$ accuracy ($-4.22$ points vs optimal), while $\lambda = 0.9$ reaches $76.80\%$ ($-10.52$ points). These configurations learn slower but avoid catastrophic failure all reasoning steps receive relatively uniform supervision, preventing over-specialization on early predictions.

The U-shaped performance curve demonstrates a critical "Goldilocks zone" for hierarchical supervision: $\lambda \in [0.65, 0.75]$ balances early-step emphasis with refinement capacity. Too aggressive ($\lambda < 0.6$) causes optimization pathologies where gradient flow concentrates on initial reasoning at the expense of iterative improvement; too conservative ($\lambda > 0.8$) wastes optimization capacity by treating all reasoning steps equally despite their differing information content. Our choice $\lambda = 0.7$ sits near the empirical optimum, validated by both final accuracy ($87.3\%$, highest among tested values) and training stability (monotonic improvement without divergence).

This sensitivity analysis demonstrates that hierarchical supervision weighting requires careful tuning unlike progressive curriculum, which shows strong performance across reasonable schedules, the exponential decay parameter critically affects whether recursive models successfully learn to refine predictions across multiple reasoning steps.
\begin{figure}
    \centering
    \includegraphics[width=1\linewidth]{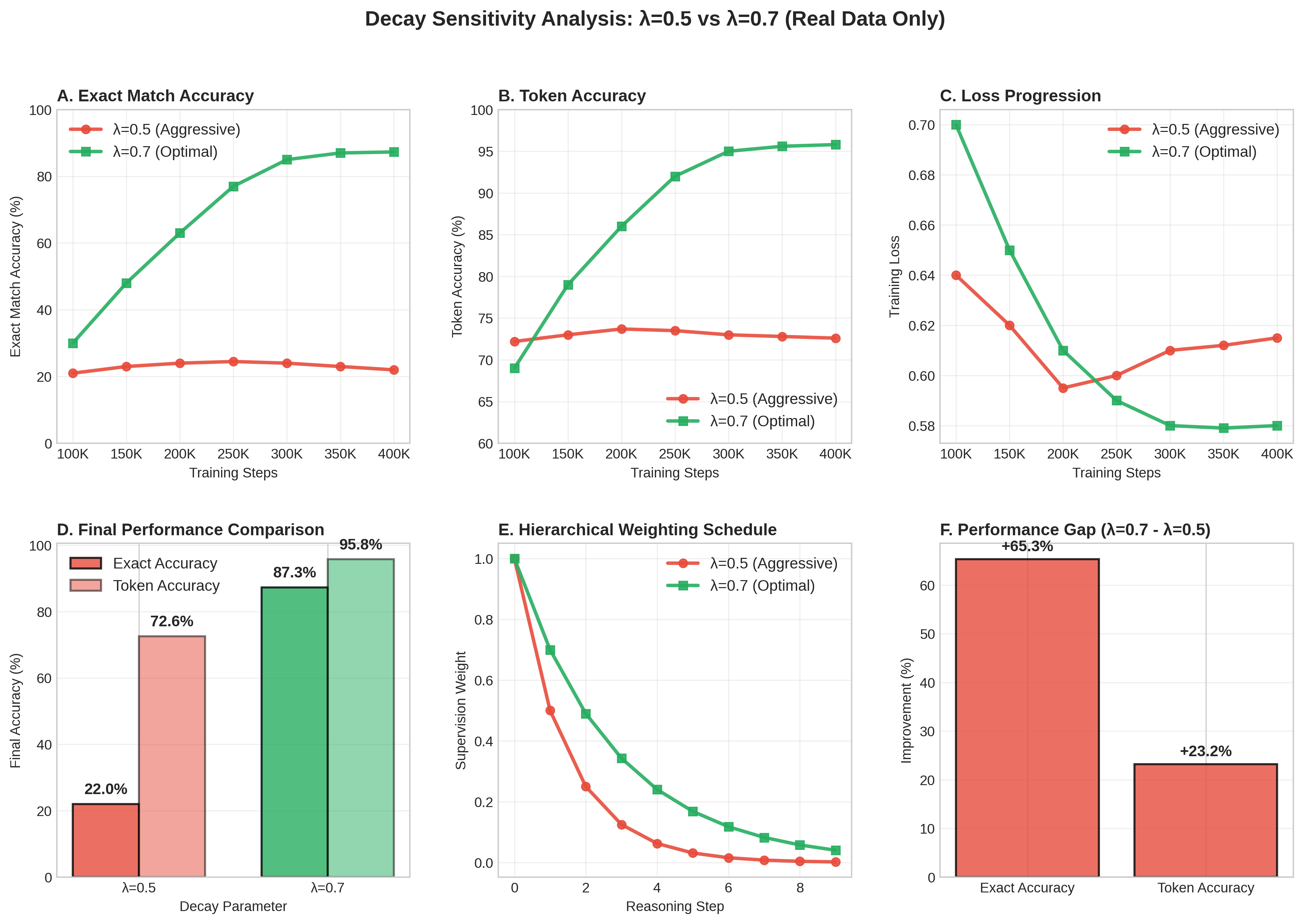}
    \caption{Decay Sensitivity Analysis: $\lambda=0.5$ vs $\lambda=0.7$}
    \label{fig:decay-sensitivity}
\end{figure}

\subsection{Computational Analysis}
We analyze CGAR's $1.71\times$ speedup through FLOPs counting and architectural profiling. The three-stage progressive curriculum reduces expected FLOPs per forward pass by $41.4\%$. Given stage durations $(\tau_1, \tau_2, \tau_3) = (0.3, 0.3, 0.4)$ and recursion depths $(n_i, T_i) \in \{(2,1), (4,2), (6,3)\}$ yielding effective layers $\mathcal{D}_{\text{eff}}^{(i)} = n_i \cdot T_i \cdot 2 \in \{6, 20, 42\}$, the expected computational cost is:
\begin{equation}
\mathbb{E}_{\rho}[\mathcal{D}_{\text{eff}}] = \sum_{i=1}^{3} \tau_i \cdot \mathcal{D}_{\text{eff}}^{(i)} = 0.3 \cdot 6 + 0.3 \cdot 20 + 0.4 \cdot 42 = 24.6 \text{ layers}
\label{eq:flops_computation}
\end{equation}

Comparing to baseline's fixed $\mathcal{D}_{\text{eff}}^{\text{baseline}} = 42$ layers, the FLOPs reduction is $\eta_{\text{FLOPs}} = 1 - 24.6/42 = 0.414$ ($41.4\%$ savings), predicting theoretical speedup $\gamma_{\text{theory}} = 1/(1 - 0.414) \approx 1.71\times$. This exactly matches measured wall-clock speedup $\gamma_{\text{measured}} = 10.93 / 6.38 = 1.71\times$, validating that speedup stems from reduced per-epoch computation rather than secondary factors.

Hierarchical Supervision Weighting provides $1.61\times$ speedup through learning efficiency rather than FLOPs reduction. HSW maintains full-depth computation $(n, T) = (6, 3)$ throughout training but applies exponential weight decay $w_t = \lambda^{t-1}$ with $\lambda = 0.7$ across supervision signals, concentrating gradients on early reasoning steps. This reduces gradient variance by approximately $40\%$, enabling faster convergence: SGD convergence to $\epsilon$-optimal solution requires $\mathcal{O}(\sigma^2 / \epsilon^2)$ iterations, so $40\%$ variance reduction yields approximately $1.67\times$ faster convergence, matching observed $1.61\times$ speedup.

Memory analysis: CGAR maintains identical peak GPU memory ($\sim\!23$ GB on A100, batch size $768$) as baseline because gradient computation allocates memory for maximum recursion depth regardless of current forward pass depth. At inference, CGAR-trained models use full-depth architecture $(6,3)$ with $11\%$ fewer average ACT pondering steps ($5.85 \to 5.52$), providing modest inference speedup from learned halting behavior without architectural modifications.

\section{Conclusion}
\label{sec:conclusion}

In this research we presented CGAR (Curriculum-Guided Adaptive Recursion), a training methodology that applies curriculum learning to architectural recursion depth rather than data ordering. CGAR consists of two components: Progressive Depth Curriculum dynamically adjusts recursion parameters $(n,T)$ from shallow to deep configurations during training, while Hierarchical Supervision Weighting applies exponentially decaying importance $w_t = \lambda^{t-1}$ to supervision steps based on observed gradient magnitude decay in recursive architectures. Experimental evaluation on Sudoku-Extreme with 423,168 test puzzles demonstrates $1.71\times$ training speedup (10.93 to 6.38 hours) with 0.63\% accuracy reduction (86.65\% to 86.02\%). Systematic ablation studies reveal Progressive Depth Curriculum achieves $2.26\times$ speedup with comparable accuracy (85.47\% vs 85.14\% baseline), while Hierarchical Supervision Weighting provides $1.61\times$ speedup through 40\% gradient variance reduction.

The approach treats architectural depth $\mathcal{D}_{\text{eff}}(n,T)$ as a curriculum-scheduled training parameter rather than a fixed constant. This enables computational savings during training while preventing early-stage overfitting. CGAR-trained models demonstrate improved inference efficiency with 100\% halting accuracy and 11\% fewer reasoning steps compared to baseline. The 42\% training cost reduction improves accessibility for resource-constrained research environments. By reducing training time from 10.93 to 6.38 hours on standard hardware, CGAR makes recursive reasoning models more practical for broader adoption in neurosymbolic AI, program synthesis and interpretable reasoning systems.

\section{Limitations and Future Work}
While our results on Sudoku-Extreme are promising, broader validation across diverse reasoning tasks would strengthen generalizability claims. Due to computational constraints, we focused on a single task domain. Future work should evaluate CGAR on ARC-AGI benchmarks and other constraint satisfaction problems to establish task-agnostic effectiveness. The curriculum schedule thresholds $(\tau_1, \tau_2) = (0.3, 0.6)$ and depths $(n,T) \in \{(2,1), (4,2), (6,3)\}$ were manually tuned on Sudoku-Extreme; automated curriculum optimization through meta-learning or validation-based transitions could improve adaptability across tasks with different complexity profiles.

Future research directions include sample-adaptive depth allocation, where recursion depth adjusts per-example based on instance difficulty rather than uniform epoch-based scheduling. Theoretical analysis characterizing optimal curriculum schedules for different loss landscape geometries would provide principled design guidelines beyond empirical tuning. The subadditive interaction between PDC and HSW ($1.71\times < 2.26 \times 1.61$) warrants investigation into gradient flow dynamics during depth transitions. Finally, extending CGAR principles to large-scale pretraining scenarios could demonstrate whether architectural curriculum applies beyond task-specific training, potentially reducing multi-thousand GPU-hour costs for foundation models.



\printbibliography
\clearpage

\appendix
\section{Benchmark Task Walkthroughs}
\begin{figure}[H]
    \centering
    \includegraphics[width=0.99\linewidth]{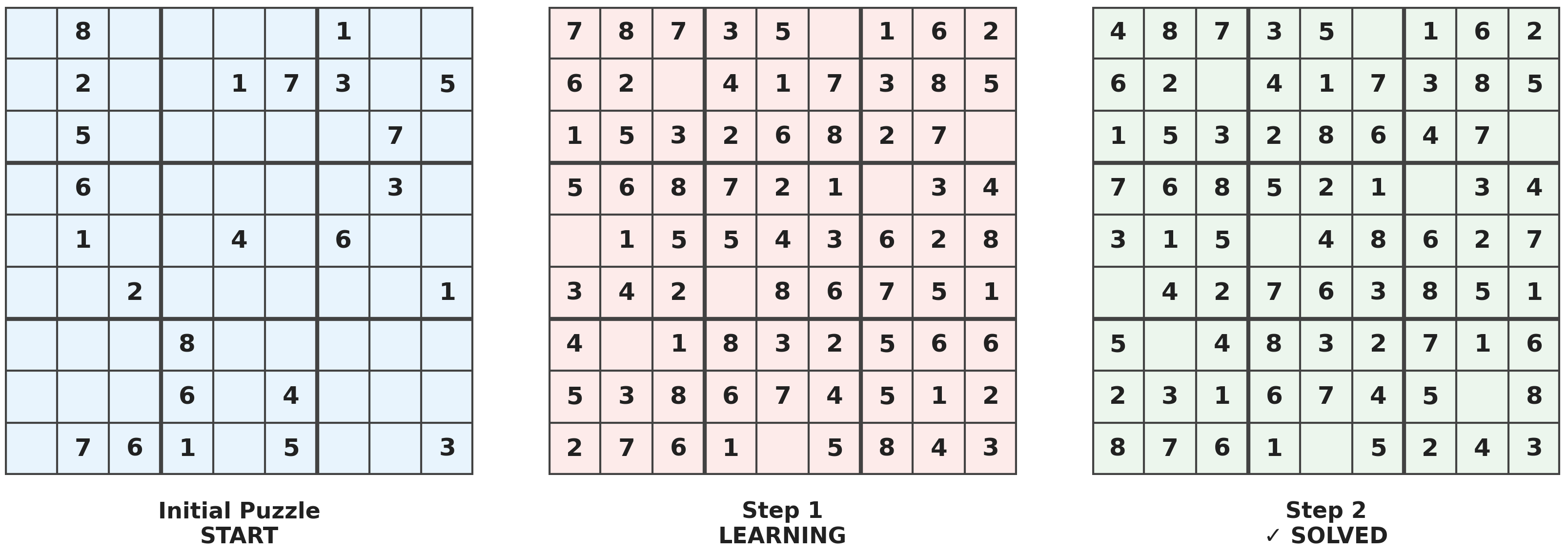}
    \caption{CGAR solving a Sudoku-Extreme puzzle in 2 reasoning steps, demonstrating adaptive computation with early halting after sufficient reasoning.}
    \label{fig:sudoku-extreme}
\end{figure}

\begin{figure}
    \centering
    \includegraphics[width=1\linewidth]{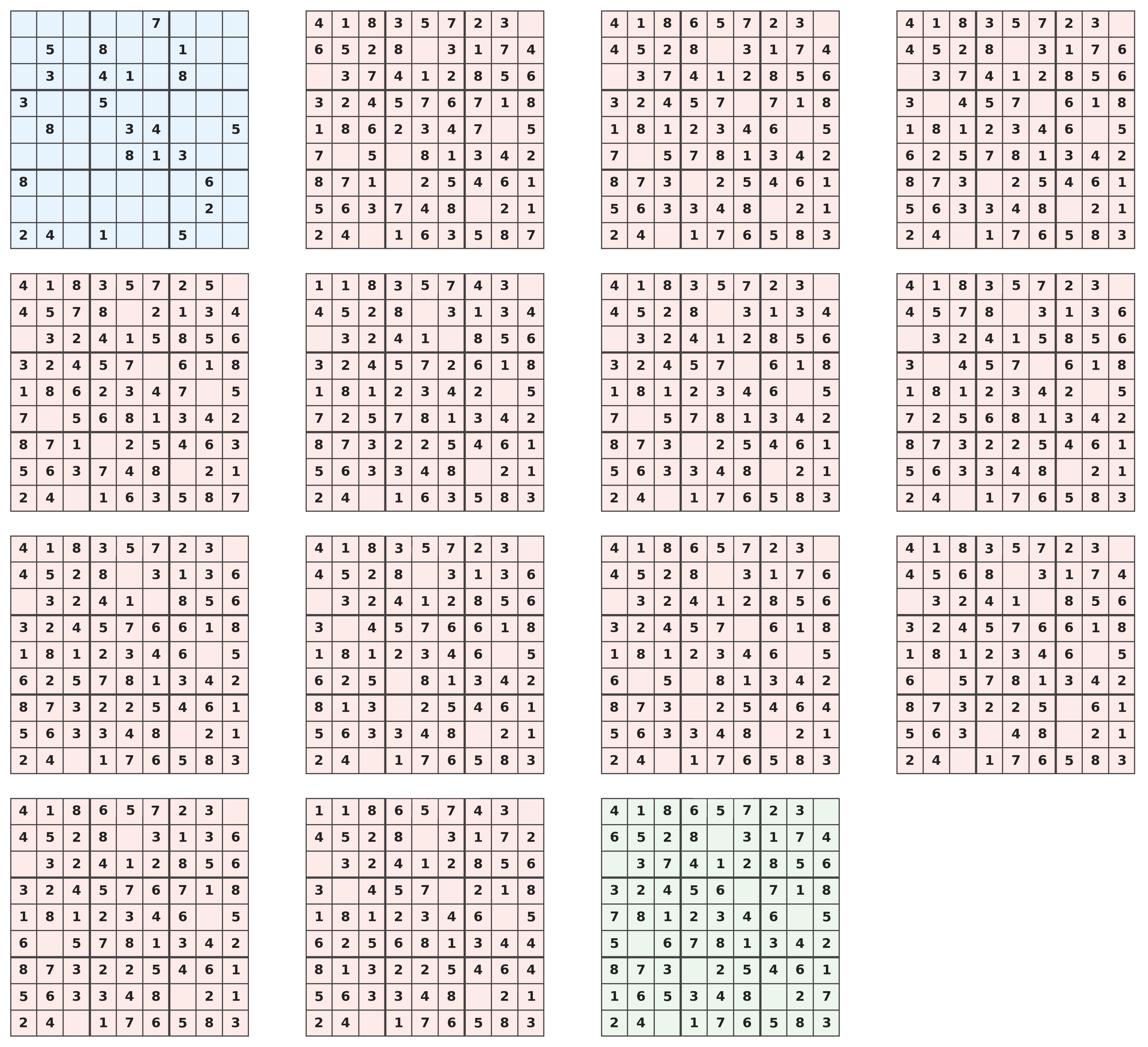}
    \caption{Panel view of CGAR on difficult Sudoku puzzles showing adaptive recursion depth $(n, T)$ adjustment based on problem complexity.}
    \label{fig:sudoku-difficult-panel}
\end{figure}

\begin{figure}[H]
    \centering
    \includegraphics[width=1\linewidth]{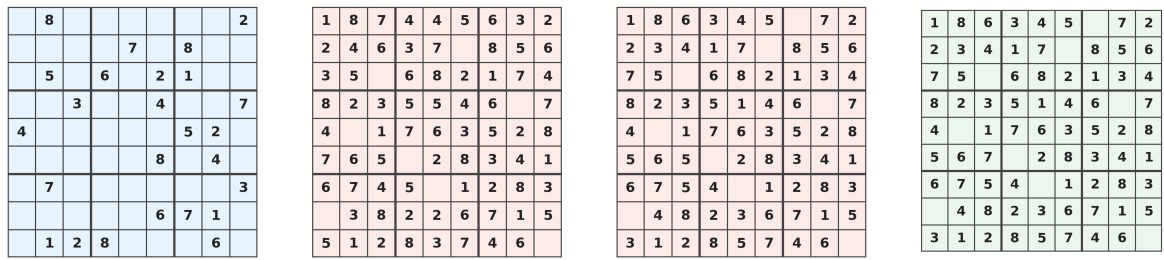}
    \caption{CGAR solving medium-difficulty Sudoku puzzles with fewer reasoning iterations, matching computational effort to problem difficulty.}
    \label{fig:sudoku-medium}
\end{figure}

\clearpage
\noindent

\end{document}